\documentclass{article}

\usepackage[final,nonatbib]{neurips_2019}

\usepackage[utf8]{inputenc} 
\usepackage[T1]{fontenc}    
\usepackage{hyperref}       
\usepackage{url}            
\usepackage{booktabs}       
\usepackage{amsfonts}       
\usepackage{nicefrac}       
\usepackage{microtype}      

\usepackage{bm}
\usepackage{multirow}
\usepackage{amsmath}
\usepackage{amssymb}
\usepackage{amsthm}
\usepackage{color}
\usepackage{times}
\usepackage{graphicx} 
\usepackage{subfigure} 
\usepackage[rightcaption]{sidecap}
\usepackage{threeparttable}
\usepackage{algorithm}
\usepackage{algorithmic}

\title{Fused Gromov-Wasserstein Alignment\\ for Hawkes Processes}

\author{
  Dixin Luo$^{1*}$\quad\quad Hongteng Xu$^{1,2}$\thanks{Equal contribution}\quad\quad  Lawrence Carin$^{1}$ \\
  $^1$Duke University \quad\quad $^2$Infinia ML, Inc.\\
  \texttt{\{dixin.luo,~hongteng.xu\}@duke.edu} \\
}

\begin{document}

\maketitle

\begin{abstract}
We propose a novel fused Gromov-Wasserstein alignment method to jointly learn the Hawkes processes in different event spaces, and align their event types.
Given two Hawkes processes, we use fused Gromov-Wasserstein discrepancy to measure their dissimilarity, which considers both the Wasserstein discrepancy based on their base intensities and the Gromov-Wasserstein discrepancy based on their infectivity matrices.
Accordingly, the learned optimal transport reflects the correspondence between the event types of these two Hawkes processes.
The Hawkes processes and their optimal transport are learned jointly via maximum likelihood estimation, with a fused Gromov-Wasserstein regularizer.
Experimental results show that the proposed method works well on synthetic and real-world data.
\end{abstract}

\section{Introduction}\vspace{-8pt}
There is often a need to align real-world entities in different domains, based on their sequential behavior in continuous time, $e.g.$, linking accounts in different social networks based on behaviors within each network. 
For each domain, the entities in the domain formulate an event space and their sequential behavior can be represented as event sequences, in which each event is a tuple containing a timestamp and an event type ($i.e.$, the entity involved in the event). 
When these event sequences yield multi-dimensional point process models, the proposed problem can be reformulated as an alignment problem: learning two point processes and finding the correspondence between their event types. 

Focusing on event sequences that are modeled as Hawkes processes, we propose a novel fused Gromov-Wasserstein alignment (FGWA) method. 
As illustrated in Figure~\ref{fig:scheme}, the event sequences in each domain are modeled as a Hawkes process parametrized via a base intensity vector and an infectivity matrix. 
The base intensity captures the intrinsic expected happening rate of each event type, while the infectivity matrix describes the self- and mutually-triggering pattern between different event types.
The Wasserstein discrepancy between the two domains is formulated based on their base intensities, and their Gromov-Wasserstein discrepancy is formulated based on their infectivity matrices. 
We learn an optimal transport to minimize the fusion of these two discrepancies, $i.e.$, the fused Gromov-Wasserstein discrepancy~\cite{vayer2018fused}. 
The learned optimal transport are used to regularize updating of the Hawkes processes. 
After several iterations, we jointly derive the two Hawkes processes and the optimal transport, indicating the correspondence between their event types. 
As shown in Fig.~\ref{fig1}(b-e), compared with its competitors our FGWA method learns the optimal transport matrix with the highest certainty --- each row just contains one nonzero element.

\section{Proposed Alignment Method}\vspace{-8pt}
A temporal point process with $C$ event types can be represented as a counting process $N(t)=\{N_c(t)\}_{c=1}^C$, where each $N_c(t)$ is the number of type-$c$ events happening at or before time $t$. 
The event sequences of the point process are denoted $\mathcal{S}=\{\bm{s}_n=(t_i^n, c_i^n)_{i=1}^{I_n}\}_{n=1}^{N}$, where $N$ is the number of sequences, $I_n$ is the number of events in $\bm{s}_n$, with $t_i^n\in [0, T]$ and $c_i^n\in\mathcal{C}=\{1,...,C\}$ representing respectively the time-stamp and the event type of the $i$-th event. 
Point processes are characterized by their intensity functions $\{\lambda_c(t)\}_{c=1}^\mathcal{C}$, where $\lambda_c(t)=\mathbb{E}[dN_c(t)|\mathcal{H}^\mathcal{C}(t)]/dt$ represents the expected instantaneous happening rate of type-$c$ events given the history $\mathcal{H}^\mathcal{C}(t)=\{(t_i, c_i)|t_i<t, c_i\in \mathcal{C}\}$. 
As a special kind of point process, the Hawkes process~\cite{hawkes1971spectra} has a particular form of intensity~\cite{luo2015multi,zhou2013learning}:
\begin{eqnarray}\label{eq:intensity}
\begin{aligned}
\lambda_c(t)=\mu_c+\sideset{}{_{i:t_i<t}}\sum\phi_{cc_i}(t-t_i),~\mbox{for}~c\in\mathcal{C}.
\end{aligned}
\end{eqnarray}
Here, $\mu_c$ is the base intensity, independent of history, capturing the intrinsic happening rate of the type-$c$ event, and $\phi_{cc'}(t)$ is the impact function measuring the infectivity of the type-$c'$ event to the type-$c$ event type, over time. 
Generally, we can parameterize each impact function by a predefined base function, $i.e.$, $\phi_{cc'}(t)=a_{cc'}g(t)$, where $g(t)$ is an exponential function and $a_{cc'}$ is a learnable coefficient. 
Therefore, we denote an event sequence yielding to a Hawkes process as $\bm{s}\sim \mbox{HP}(\bm{\mu}, \bm{A})$, with basic intensity $\bm{\mu}\in\mathbb{R}^{C}$ and infectivity matrix $\bm{A}=[a_{cc'}]\mathbb{R}^{C\times C}$. 
Given a set of event sequences $\mathcal{S}$, we can learn a Hawkes processes via maximum likelihood estimation. 
The likelihood of $\mathcal{S}$ is
\begin{eqnarray}\label{eq:loglike}
\begin{aligned}
L(\mathcal{S};\bm{\mu}, \bm{A})=\sideset{}{_{n}}\prod p(\bm{s}_n; \bm{\mu}, \bm{A})=\sideset{}{_n}\prod\sideset{}{_{i=1}^{I_n}}\prod\lambda_{c_i^n}(t_i^n)\exp\Bigl(-\sideset{}{_{c=1}^{C}}\sum\int_{0}^{T_n}\lambda_c(s)ds\Bigr).
\end{aligned}
\end{eqnarray}

\begin{figure}[t]
    \centering
    \subfigure[\tiny{Proposed method}]{
    \includegraphics[height=3.5cm]{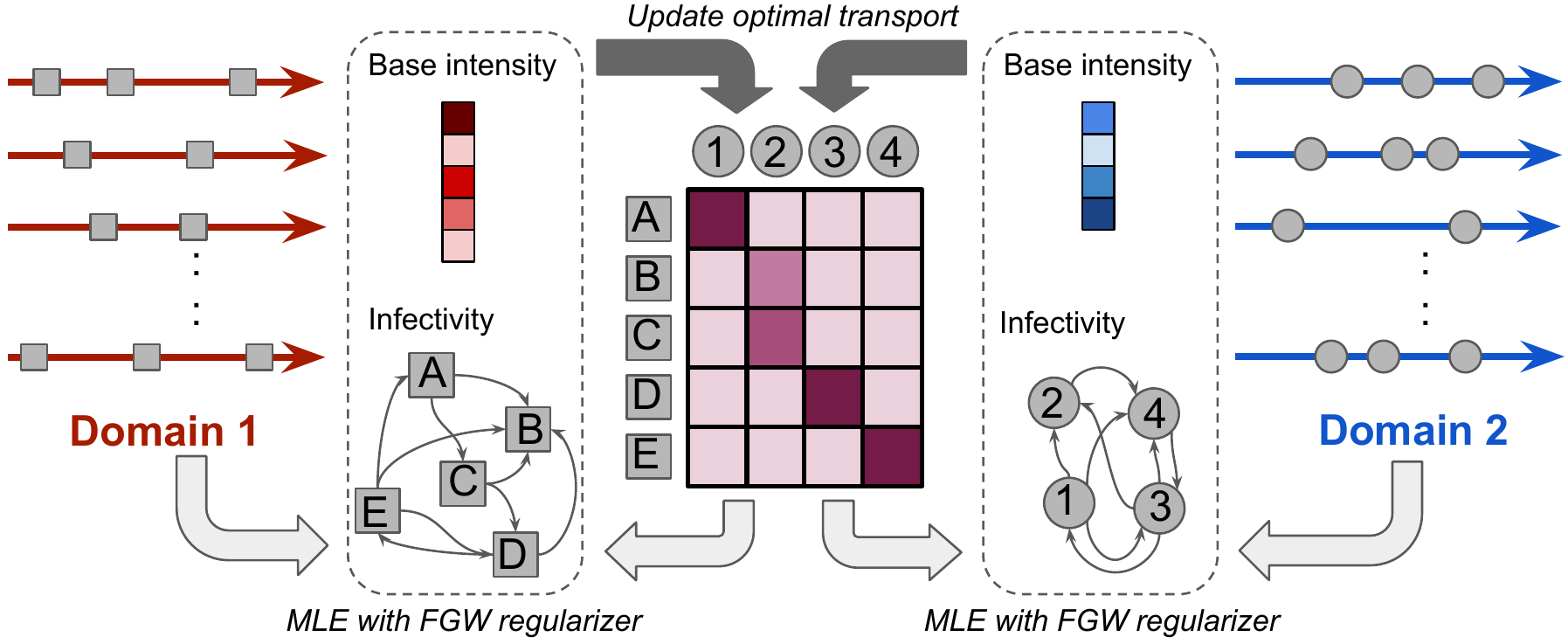}\label{fig:scheme}
    }
    \begin{minipage}[b]{0.14\linewidth}
    \centering
    \subfigure[\tiny{Empirical}]{
    \includegraphics[width=1\linewidth]{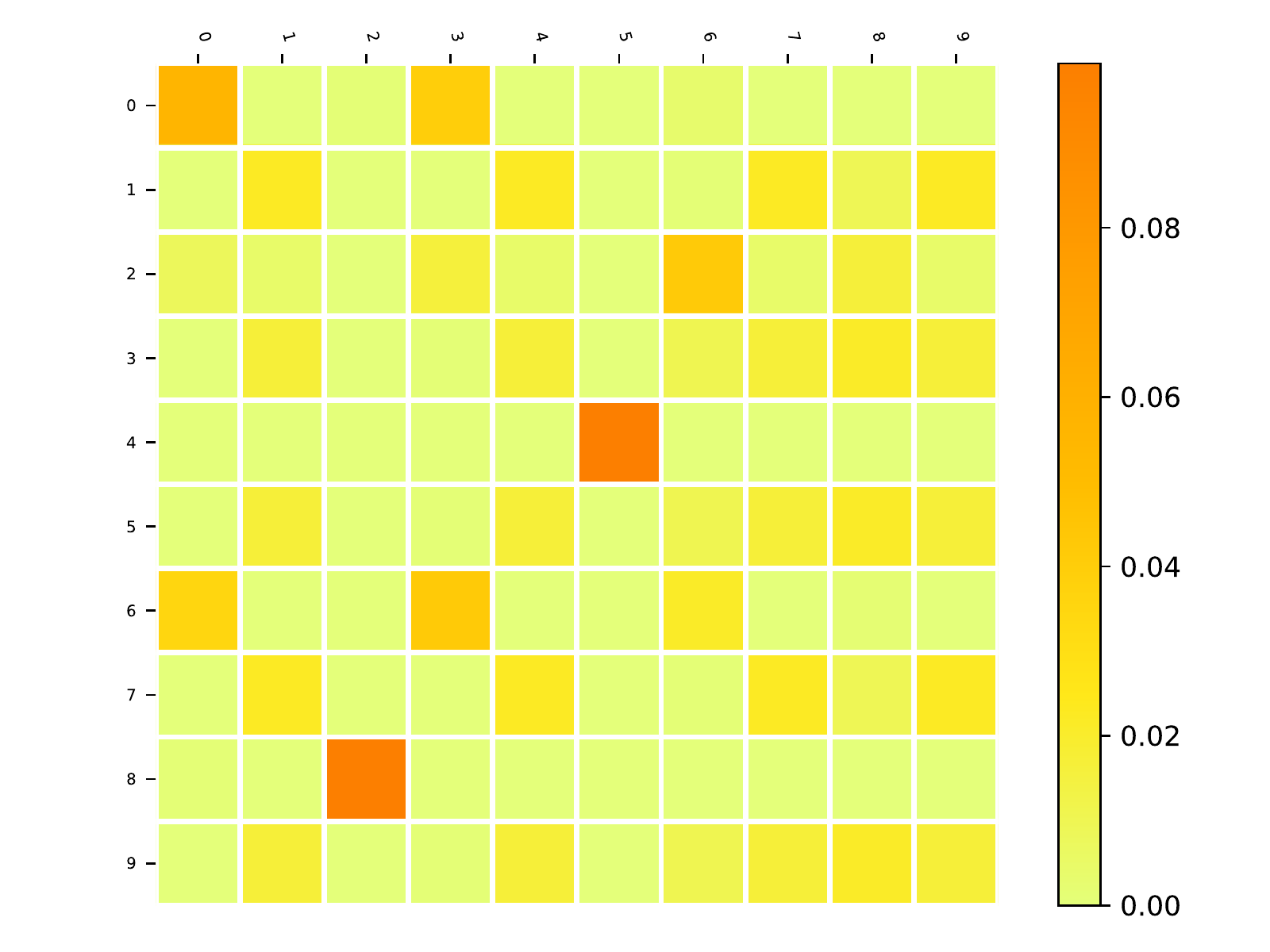}
    }\\
    \vspace{-8pt}
    \subfigure[\tiny{HP-WD}]{
    \includegraphics[width=1\linewidth]{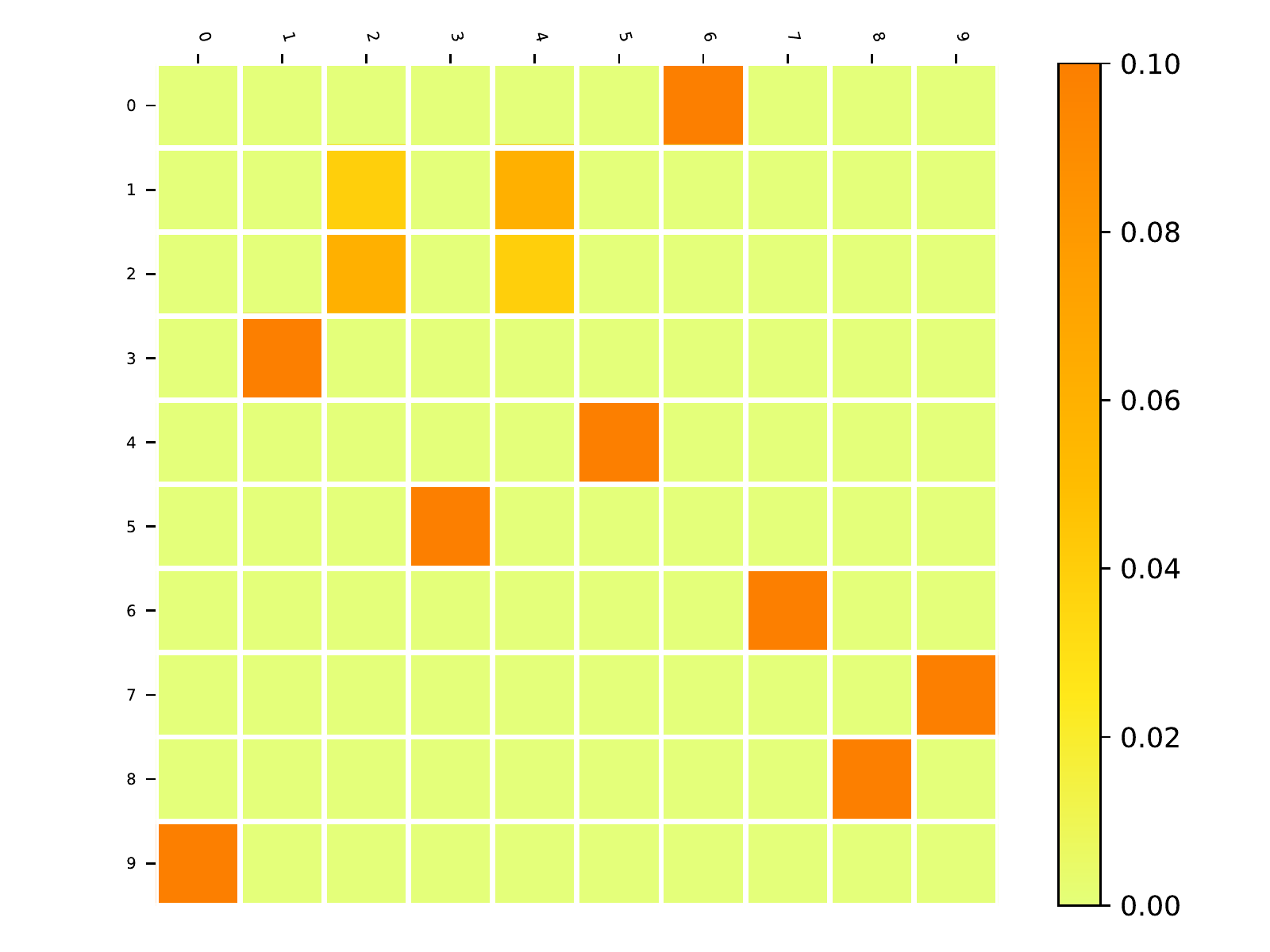}
    \vspace{-8pt}
    }
    \end{minipage}
    \begin{minipage}[b]{0.14\linewidth}
    \subfigure[\tiny{HP-GWD}]{
    \includegraphics[width=1\linewidth]{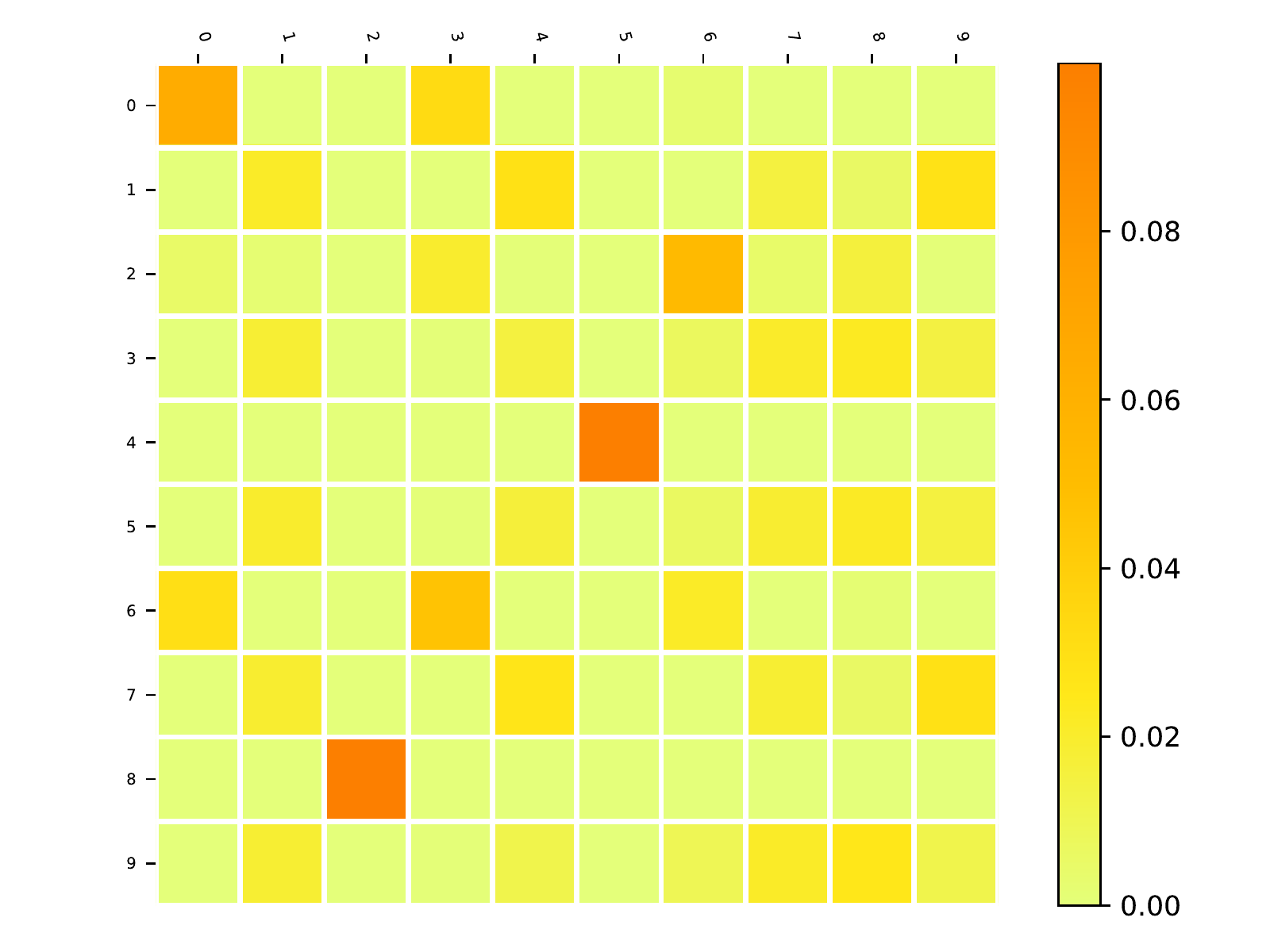}
    }\\
    \vspace{-8pt}
    \subfigure[\tiny{\textbf{FGWA (Ours)}}]{
    \includegraphics[width=1\linewidth]{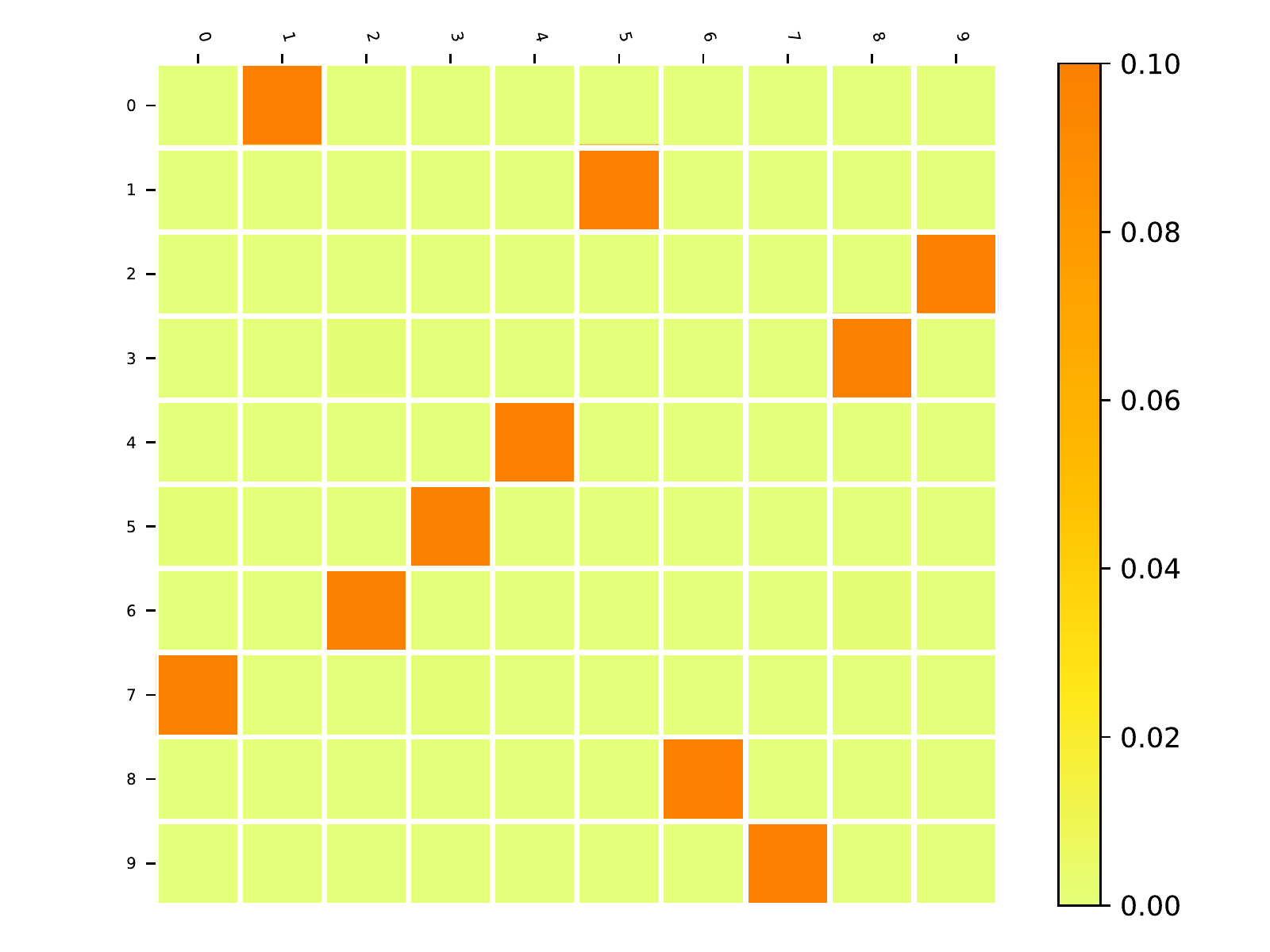}
    \vspace{-8pt}
    }
    \end{minipage}
    \vspace{-8pt}
    \caption{(a) An illustration of our method. (b-e) Comparisons on synthetic data.}
    \label{fig1}
\end{figure}

The base intensity and the infectivity matrix provide, respectively, the feature of each event type and the relationship among different event types. 
These two kinds of information can be applied to measure the similarity between different event types in a framework of fused Gromov-Wasserstein discrepancy~\cite{vayer2018fused}. 
In particular, fused Gromov-Wasserstein discrepancy is a combination of traditional Wasserstein discrepancy (WD)~\cite{villani2008optimal} and Gromov-Wasserstein discrepancy (GWD)~\cite{peyre2016gromov}.  
Focusing on the alignment of Hawkes processes, the proposed fused Gromov-Wasserstein discrepancy can be used as a regularizer when learning the Hawkes process models. 
Suppose that we have two sets of event sequences corresponding to source and target Hawkes processes, $i.e.$, $\mathcal{S}_s\sim \mbox{HP}(\bm{\mu}_s,\bm{A}_s)$ and $\mathcal{S}_t\sim \mbox{HP}(\bm{\mu}_t,\bm{A}_t)$, where $\bm{\mu}_k=[\mu_i^k]\in\mathbb{R}^{C_k}$ and $\bm{A}_k=[a_{ij}^k]\in\mathbb{R}^{C_k\times C_k}$ for $k=s$ and $t$. 
We learn these two Hawkes processes and align their event types via maximum likelihood estimation with a fused Gromov-Wasserstein regularizer:
\begin{eqnarray}\label{eq:mle-fgw}
\min_{\{\bm{\mu}_k,\bm{A}_k\geq\bm{0}\}_{k=s,t}}&\underbrace{-\sideset{}{_{k=s,t}}\sum \log L(\mathcal{S}_k;\bm{\mu}_k,\bm{A}_k)}_{\text{negative log-likelihood}}+\gamma\underbrace{ d_{fgw}^2(\bm{u}_s,\bm{u}_t;\bm{\mu}_s,\bm{\mu}_t,\bm{A}_s,\bm{A}_t)}_{\text{Fused Gromov-Wasserstein discrepancy}},
\end{eqnarray}
where $\bm{u}_s$ and $\bm{u}_t$ represent the empirical distribution of the event type in the source and target domain, respectively. 
These are estimated via the histograms of the counts of events according to $\mathcal{S}_s$ and $\mathcal{S}_t$. 
The hyperparameter $\gamma$ controls the significance of the proposed fused Gromov-Wasserstein regularizer. 
$d_{fgw}^2(\bm{u}_s,\bm{u}_t;\bm{\mu}_s,\bm{\mu}_t,\bm{A}_s,\bm{A}_t)$ is the discretized version of fused Gromov-Wasserstein discrepancy based on the Hawkes process parameters:
\begin{eqnarray}\label{eq:dfgw}
\begin{aligned}
d_{fgw}^2
&=\sideset{}{_{\bm{T}\in\Pi(\bm{u}_s,\bm{u}_t)}}\min(1-\alpha)\underbrace{\sideset{}{_{i,j}}\sum L(\mu_{i}^s,\mu_{j}^t)T_{ij}}_{\text{Wasserstein term}}+\alpha\underbrace{\sideset{}{_{i,j,i',j'}}\sum L(a_{ij}^s,a_{i'j'}^t)T_{ii'}T_{jj'}}_{\text{Gromov-Wasserstein term}}\\
&=\sideset{}{_{\bm{T}\in\Pi(\bm{u}_s,\bm{u}_t)}}\min\langle (1-\alpha)\bm{L}_{\mu}+\alpha\bm{L}_A(\bm{T}), \bm{T}\rangle,
\end{aligned}
\end{eqnarray}
where $L(a,b)=|a-b|^2$ is a mean-square-error (MSE) loss, and $\langle\cdot,\cdot\rangle$ represents the matrix inner product. 
Accordingly, $\bm{L}_{\mu}=[L(\mu_{i}^s,\mu_{j}^t)]\in\mathbb{R}^{C_s\times C_t}$ and $\bm{L}_{A}(\bm{T})=[L_{jj'}]$, whose element $L_{jj'}=\sum_{i, i'}L(a_{ij}^s,a_{i'j'}^t)T_{ii'}$; and  
$\Pi(\bm{u}_s,\bm{u}_t)=\{\bm{T}\geq \bm{0}|\bm{T}\bm{1}_{C_t}=\bm{u}_s,~\bm{T}^{\top}\bm{1}_{C_s}=\bm{u}_t\}$, where $\bm{1}_D$ represents a $D$-dimensional all-one vector. 
$\alpha$ controls the balance between the Wasserstein term and the Gromov-Wasserstein term. 
The Wasserstein discrepancy compares the event types of the two Hawkes processes in an absolute way while the Gromov-Wasserstein discrepancy compares their event types in a relational way. 
Taking them into account, the final optimal transport represents the joint distribution of the event types in different Hawkes processes. 
As shown in Figure~\ref{fig:scheme}, the pairs of event types with high probability indicate the correspondence between the event types. 
The learned optimal transport fills the gap between the source and the target Hawkes processes, and the models can be learned jointly under the guidance of the optimal transport.

\section{Learning Algorithm}\vspace{-8pt}
We solve (\ref{eq:mle-fgw}) effectively based on an alternating optimization strategy. 
In each iteration, given the current Hawkes process models, we update the optimal transport between them, and then the Hawkes processes are updated based on the learned optimal transport. 

\textbf{Updating Hawkes processes}
In the $n$-th iteration, given the optimal transport learned in the previous iteration, $i.e.$, $\bm{T}^{(n-1)}$, we update the Hawkes process models by
\begin{eqnarray}\label{eq:updateHP}
\sideset{}{_{\{\bm{\mu}_k,\bm{A}_k\geq\bm{0}\}_{k=s,t}}}\min-\sideset{}{_{k=s,t}}\sum \log L(\mathcal{S}_k;\bm{\mu}_k,\bm{A}_k)+\gamma\langle (1-\alpha)\bm{L}_{\mu}+\alpha\bm{L}_A(\bm{T}^{(n-1)}), \bm{T}^{(n-1)}\rangle,
\end{eqnarray}
This problem can be solved effectively via stochastic gradient descent (SGD)~\cite{mei2017neural}. 
We randomly select a batch of events and their historical events, and calculate the gradients of the base intensities and the infectivity matrices related to the event types appearing in the batch. 
After the parameters are updated via gradient descent, they are projected into the nonnegative space to match the constraints in (\ref{eq:mle-fgw}). 

\textbf{Updating optimal transport}
Given updated Hawkes processes, we further update the optimal transport by solving the following optimization problem:
\begin{eqnarray}\label{eq:updateOT}
\begin{aligned}
\sideset{}{_{\bm{T}\in\Pi(\bm{u}_s,\bm{u}_t)}}\min \langle (1-\alpha) \bm{L}_{\mu}^{(n)} + \alpha\bm{L}_{A}^{(n)}(\bm{T}), \bm{T}\rangle,
\end{aligned}
\end{eqnarray}
where $\bm{L}_{\mu}^{(n)}$ and $\bm{L}_{A}^{(n)}(\bm{T})$ are calculated based on the updated base intensities and infectivity matrices.
Inspired by the work in~\cite{peyre2016gromov,xu2019gromov}, we apply a proximal gradient method to solve (\ref{eq:updateOT}) iteratively. 
Given current optimal transport $\bm{T}^{(n)}$, we add a proximal term as the regularizer of (\ref{eq:updateOT}):
\begin{eqnarray}\label{eq:proximal}
\begin{aligned}
\sideset{}{_{\bm{T}\in\Pi(\bm{u}_s,\bm{u}_t)}}\min &\langle \alpha\bm{L}_{\mu}^{(n)}+(1-\alpha)\bm{L}_{A}^{(n)}(\bm{T}), \bm{T}\rangle+\tau \mbox{KL}(\bm{T}\lVert\bm{T}^{(n)}),
\end{aligned}
\end{eqnarray}
where $\mbox{KL}(\bm{T}\lVert\bm{T}^{(n)})=\sum_{i,j}T_{ij}\log({T_{ij}}/{T_{ij}^{(m)}}) - T_{ij} + T_{ij}^{(n)}$ is the Kullback-Leibler (KL) divergence. 
Applying the proximal gradient method, (\ref{eq:proximal}) is solved iteratively, and each iteration corresponds to solving the following problem via Sinkhorn iterations~\cite{xu2019gromov}.

When updating the Hawkes processes, sub-problem (\ref{eq:updateHP}) is convex and can be solved with a high convergence rate.
When updating the optimal transport, the proposed algorithm is a special case of successive upper-bound minimization (SUM)~\cite{razaviyayn2013unified}, whose global convergence is guaranteed.
Applying SGD, we solve (\ref{eq:updateHP}) with computational complexity $\mathcal{O}(BK)$, where $B$ is the size of batch ($i.e.$, the number of selected events), and $K$ is the length of each event's history. 
Because in general $B\ll \sum_n I_n$ and $K\ll \max I_n$, the updating of the Hawkes processes scales well. 
The complexity of updating the optimal transport is $\mathcal{O}(C^3)$.
Both these two steps can be done in parallel on GPUs. 

\section{Experimental Results}\vspace{-8pt}
To demonstrate the feasibility and the effectiveness of the proposed alignment method (\textbf{FGWA}), we consider both synthetic and real-world data. 
In the following experiments, we set $\alpha=0.8$, which balances the influence of Wasserstein discrepancy and that of Gromov-Wasserstein discrepancy. 
We compare our method with the following baselines: 
1) aligning event types according to their empirical distributions $\bm{u}_s$ and $\bm{u}_t$ directly (\textbf{Empirical}); 
2) aligning Hawkes process purely based on Wasserstein discrepancy, $i.e.$, $\alpha=0$ (\textbf{HP-WD}); and 
3) aligning Hawkes process purely based on Gromov-Wasserstein discrepancy, $i.e.$, $\alpha=1$ (\textbf{HP-GWD}). 
Given the real correspondence $\bm{T}$ and the optimal transport $\widehat{\bm{T}}$, we evaluate various methods based on the following three measurements: 
i) \textit{Top-$K$ alignment accuracy} $\text{Acc}=\langle\bm{T},\text{top}_{K}(\widehat{\bm{T}})\rangle/C$, where $\text{top}_{K}(\widehat{\bm{T}})$ converts each row of $\widehat{\bm{T}}$ to binary vector, whose nonzero elements corresponds to the maximum $K$ values of the row.
ii) \textit{Cosine similarity} $\text{Sim}=\langle\bm{T},\widehat{\bm{T}}\rangle/(\|\bm{T}\|_F\|\widehat{\bm{T}}\|_F)$.
iii) \textit{Entropy} $H=-\langle\widehat{\bm{T}},\log\widehat{\bm{T}}\rangle$. When the real correspondence is bijective, this measurement reflects the uncertainty of the learned correspondence.

\begin{table*}[t]
\caption{Comparisons for various methods on synthetic and real-world data.}
\centering
\small{
\begin{tabular}{c|
c@{\hspace{5pt}}c@{\hspace{5pt}}c|
c@{\hspace{5pt}}c@{\hspace{5pt}}c|
c@{\hspace{5pt}}c@{\hspace{5pt}}c|
c@{\hspace{5pt}}c@{\hspace{5pt}}c} 
\hline\hline
Method &
\multicolumn{3}{c|}{Empirical} &
\multicolumn{3}{c|}{HP-WD} &
\multicolumn{3}{c|}{HP-GWD} &
\multicolumn{3}{c}{FGWA}\\ 
\hline
Synthetic
&$\text{Acc-}1$ &$\text{Sim}$ &$H$
&$\text{Acc-}1$ &$\text{Sim}$ &$H$
&$\text{Acc-}1$ &$\text{Sim}$ &$H$
&$\text{Acc-}1$ &$\text{Sim}$ &$H$\\ \hline
$C$=10 
&0.41    &0.45    &3.23
&0.43    &0.48    &\textbf{2.30}
&0.49    &0.45    &3.20
&\textbf{0.69}    &\textbf{0.50}    &\textbf{2.30}\\ 
$C$=50 
&0.12    &0.08    &7.63
&0.19    &\textbf{0.12}    &4.75
&0.18    &0.09    &7.62
&\textbf{0.22}    &\textbf{0.12}    &\textbf{4.60}\\ 
$C$=100 
&0.03    &0.02   &12.43
&0.06    &0.05   &9.66
&0.06    &0.05   &12.42
&\textbf{0.11}    &\textbf{0.06}   &\textbf{9.60}\\ 
\hline\hline
Real-world
&$\text{Acc-}K$ &$\text{Sim}$ &--
&$\text{Acc-}K$ &$\text{Sim}$ &--
&$\text{Acc-}K$ &$\text{Sim}$ &--
&$\text{Acc-}K$ &$\text{Sim}$ &--\\ \hline
MIMIC-III
&0.196 &0.251 &--
&0.332 &0.469 &--
&0.314 &0.336 &--
&\textbf{0.464} &\textbf{0.471} &--\\
MC3
&0.081 &0.061 &--
&0.177 &0.099 &--
&0.129 &0.102 &--
&\textbf{0.253} &\textbf{0.106} &--\\ 
\hline\hline
\end{tabular}\label{tab:exp}
}
\end{table*}

\textbf{Synthetic data}
The synthetic event sequences are generated via the following method: For the source Hawkes process with $C_s$ event types, we generate $\bm{\mu}_s=[\mu_i^s]$ with $\mu_i^s\sim \text{Uniform}[0, {1}/{C_s}]$ and $\bm{A}_s=[a_{ij}^s]$ with $a_{ij}^s\sim \text{Uniform}[0, {1}/{C_s^2}]$. 
Given a predefined correspondence matrix $\bm{T}$, the parameters of the target Hawkes process are $\bm{\mu}_t = \bm{T}^{\top}\bm{\mu}_s$ and $\bm{A}_t = \bm{T}^{\top}\bm{A}_s\bm{T}$. 
Accordingly, the source and the target event sequences are generated based on Ogata's thinning algorithm~\cite{ogata1981lewis}.
We keep $C_s=C_t$ and set them from $\{10, 50, 100\}$. 
For both the source and target Hawkes process, we simulate $C$ ($C=C_s=C_t$) event sequences with length $T=C^2$, and set the decay function as an exponential function, $i.e.$, $g(t)=\exp(-t)$. 
We consider $30$ trials, and calculate the average results. 
Because the real correspondence in each trial is a bijective function, we consider the top-$1$ alignment accuracy in this experiment. 
Table~\ref{tab:exp} shows the results of various methods. 
The proposed FGWA method outperforms its competitors in most situations. 
Considering the fused Gromov-Wasserstein regularizer is beneficial for our alignment task indeed. 
The entropy of our optimal transport matrix is often smaller than those of other methods, which means that the correspondence we have learned has high certainty. 
The optimal transport matrices shown in Figure~\ref{fig1}(b-e) further demonstrates that our FGWA method has the highest certainty.

\textbf{Real-world data}
We further test the proposed method on two real-world datasets: the MIMIC-III dataset~\cite{johnson2016mimic} and the call-network used in the Mini-Challenge 3 (MC3) of VAST Challenge 2018 \url{http://vacommunity.org/VAST+Challenge+2018+MC3}. 
The MIMIC-III records 18,756 patient admission sequences. 
Each admission is an event in the sequence, containing a pair of diagnose ICD code and procedure ICD code. 
The dataset contains 56 diagnoses and 25 procedures. 
According to the coherency of the diagnoses and the procedures in the observed admission sequences, we obtain the correspondence between them. 
The call-network we used records the phone calls among a company's employees in continuous time domain, which contains 2,507 callers and 2,481 responders.
The pairs of callers and responders appearing in the call-network indicates the correspondence between them. 
For the MIMIC-III dataset, we consider the sequences of diagnoses and those of procedures, and model them via two Hawkes processes. 
Applying various alignment methods, we try to estimate the correspondence between diagnoses and procedures. 
Similarly, for the MC3 dataset, we model the sequences of callers and those of responders via two Hawkes processes and try to estimate the correspondence between them. 
In both of these two datasets, their correspondences are not bijective. 
Therefore, we consider top-$5$ alignment accuracy for the MIMIC-III dataset and top-$50$ alignment accuracy for the MC3 dataset, respectively.
Table~\ref{tab:exp} shows that our FGWA method outperforms other methods on both datasets.

\section{Conclusions and Future Work}\vspace{-8pt}
We have proposed an alignment method for Hawkes processes based on fused Gromov-Wasserstein discrepancy, which achieves encouraging results on matching the event types of different Hawkes processes. 
The proposed method shows the potential of optimal transport techniques to the learning and the alignment of temporal point processes.
In the future, we plan to further improve the scalability of the proposed method for large-scale applications. 

\textbf{Acknowledgements}
This research was supported in part by DARPA, DOE, NIH, ONR and NSF. 

\newpage
\small{
\bibliographystyle{ieee}
\bibliography{align_hp}

\begin{thebibliography}{10}\itemsep=-1pt

\bibitem{hawkes1971spectra}
A.~G. Hawkes.
\newblock Spectra of some self-exciting and mutually exciting point processes.
\newblock {\em Biometrika}, 58(1):83--90, 1971.

\bibitem{johnson2016mimic}
A.~E. Johnson, T.~J. Pollard, L.~Shen, H.~L. Li-wei, M.~Feng, M.~Ghassemi,
  B.~Moody, P.~Szolovits, L.~A. Celi, and R.~G. Mark.
\newblock {MIMIC-III}, a freely accessible critical care database.
\newblock {\em Scientific data}, 3:160035, 2016.

\bibitem{luo2015multi}
D.~Luo, H.~Xu, Y.~Zhen, X.~Ning, H.~Zha, X.~Yang, and W.~Zhang.
\newblock Multi-task multi-dimensional {H}awkes processes for modeling event
  sequences.
\newblock In {\em IJCAI}, 2015.

\bibitem{mei2017neural}
H.~Mei and J.~M. Eisner.
\newblock The neural {H}awkes process: A neurally self-modulating multivariate
  point process.
\newblock In {\em NIPS}, 2017.

\bibitem{ogata1981lewis}
Y.~Ogata.
\newblock On {L}ewis' simulation method for point processes.
\newblock {\em IEEE Transactions on Information Theory}, 27(1):23--31, 1981.

\bibitem{peyre2016gromov}
G.~Peyr{\'e}, M.~Cuturi, and J.~Solomon.
\newblock Gromov-{W}asserstein averaging of kernel and distance matrices.
\newblock In {\em ICML}, 2016.

\bibitem{razaviyayn2013unified}
M.~Razaviyayn, M.~Hong, and Z.-Q. Luo.
\newblock A unified convergence analysis of block successive minimization
  methods for nonsmooth optimization.
\newblock {\em SIAM Journal on Optimization}, 23(2):1126--1153, 2013.

\bibitem{vayer2018fused}
T.~Vayer, L.~Chapel, R.~Flamary, R.~Tavenard, and N.~Courty.
\newblock Fused {G}romov-{W}asserstein distance for structured objects:
  theoretical foundations and mathematical properties.
\newblock {\em arXiv preprint arXiv:1811.02834}, 2018.

\bibitem{villani2008optimal}
C.~Villani.
\newblock {\em Optimal transport: {O}ld and new}, volume 338.
\newblock Springer Science \& Business Media, 2008.

\bibitem{xu2019gromov}
H.~Xu, D.~Luo, H.~Zha, and L.~Carin.
\newblock Gromov-wasserstein learning for graph matching and node embedding.
\newblock {\em arXiv preprint arXiv:1901.06003}, 2019.

\bibitem{zhou2013learning}
K.~Zhou, H.~Zha, and L.~Song.
\newblock Learning social infectivity in sparse low-rank networks using
  multi-dimensional {H}awkes processes.
\newblock In {\em AISTATS}, 2013.

\end{thebibliography}
}

\end{document}